\documentclass[]{spie}  %>>> use for US letter paper
\usepackage{amsmath,amsfonts,amssymb}
\usepackage{graphicx}
\usepackage{subcaption}
\usepackage{array}
\usepackage{amsfonts}
\usepackage{bbm}
\usepackage{booktabs}
\usepackage{siunitx}
\usepackage{float}
\usepackage{ragged2e}

\usepackage[colorlinks=true, allcolors=blue]{hyperref}

\DeclareMathOperator{\La}{\mathcal{L}}
\newcolumntype{P}[1]{>{\centering\arraybackslash}p{#1}}
\newcolumntype{M}[1]{>{\centering\arraybackslash}m{#1}}

\title{Deep Curriculum Learning in Task Space for Multi-Class Based Mammography Diagnosis}

\author[a]{Jun Luo}
\author[b]{Dooman Arefan}
\author[b,c]{Margarita Zuley}
\author[b,c]{Jules Sumkin}
\author[a,b,d,e]{Shandong Wu}
\affil[a]{Intelligent Systems Program, School of Computing and Information, University of Pittsburgh, Pittsburgh, PA, USA}
\affil[b]{Department of Radiology, University of Pittsburgh, Pittsburgh, PA, USA}
\affil[c]{Magee-Womens Hospital, University of Pittsburgh Medical Center, Pittsburgh, PA, USA}
\affil[d]{Department of Biomedical Informatics, University of Pittsburgh, Pittsburgh, PA, USA}
\affil[e]{Department of Bioengineering, University of Pittsburgh, Pittsburgh, PA, USA}

\authorinfo{Further author information: (Send correspondence to Shandong Wu)\\Shandong Wu: E-mail: wus3@upmc.edu, Telephone: 1 412 641 2567\\Jun Luo: E-mail: jul117@pitt.edu}

\begin{document} 
\maketitle

\begin{abstract}
Mammography is used as a standard screening procedure for the potential patients of breast cancer. Over the past decade, it has been shown that deep learning techniques have succeeded in reaching near-human performance in a number of tasks, and its application in mammography is one of the topics that medical researchers most concentrate on. In this work, we propose an end-to-end Curriculum Learning (CL) strategy in task space for classifying the three categories of Full-Field Digital Mammography (FFDM), namely \textit{Malignant}, \textit{Negative}, and \textit{False recall}. Specifically, our method treats this three-class classification as a ``harder'' task in terms of CL, and create an ``easier'' sub-task of classifying False recall against the combined group of Negative and Malignant. We introduce a loss scheduler to dynamically weight the contribution of the losses from the two tasks throughout the entire training process. We conduct experiments on an FFDM datasets of 1,709 images using 5-fold cross validation. The results show that our curriculum learning strategy can boost the performance for classifying the three categories of FFDM compared to the baseline strategies for model training.
\end{abstract}


\section{INTRODUCTION}
\label{sec:intro}  % \label{} allows reference to this section

Breast cancer is the second leading cause of cancer death in women. As of 2021, breast cancer takes up 12\% of new annual cancer cases globally\cite{breastcanceror2021}. Full-Field Digital Mammography (FFDM), also known as the digital mammography, is a mammography screening techniques that enables the manipulation of images in such a way that it is more convenient for the radiologists to evaluate the areas of concerns with computer-aided detection (CAD) systems. Meanwhile, over the past decade, it has been shown that deep learning techniques have succeeded in reaching near-human performance in a number of tasks\cite{simonyan2014very,luo2021medical,jimenez2019medical}, and its application in mammography is one of the topics that medical researchers most concentrate on\cite{aboutalib2018deep}.

\begin{figure}
    \centering
    \hspace*{\fill}%
    \begin{subfigure}[b]{0.2\textwidth}
        \centering
        \includegraphics[width=\textwidth]{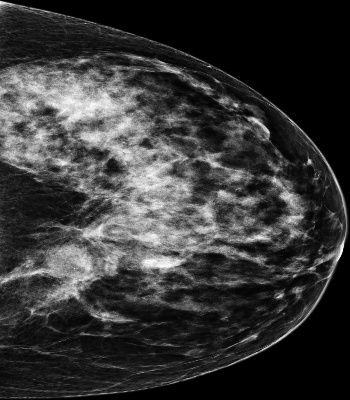}
        \caption{Malignant}
        \label{malignant}
    \end{subfigure}
    % \hfill
    \hspace*{\fill}%
    \begin{subfigure}[b]{0.2\textwidth}
        \centering
        \includegraphics[width=\textwidth]{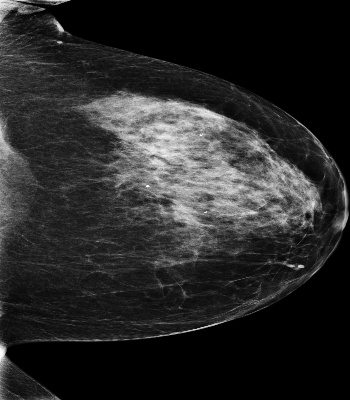}
        \caption{False recall}
        \label{falserecall}
    \end{subfigure}
    % \hfill
    \hspace*{\fill}%
    \begin{subfigure}[b]{0.2\textwidth}
        \centering
        \includegraphics[width=\textwidth]{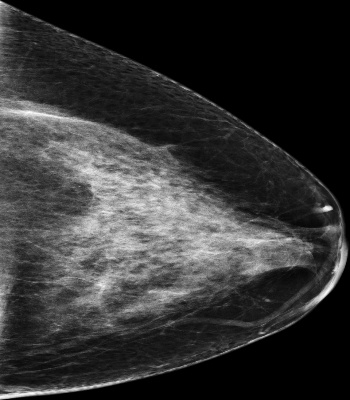}
        \caption{Negative}
        \label{negative}
    \end{subfigure}
    \hspace*{\fill}%
        \caption{Example Full-Field Digital Mammography images from the three categories of our dataset.}
        \label{ffdmexamples}
\end{figure}

Curriculum learning\cite{bengio2009curriculum} (CL), as a major machine learning regime, is a learning strategy where an ``easy-to-hard'' curriculum is designed to train the model. CL is particularly intriguing in medical fields, since medical knowledge can serve as an important contributing factor to the definition of ``easy'' and ``hard'' in terms of CL\cite{luo2021medical,jimenez2019medical}. Such a curriculum can be designed with respect to both the input space and the output (task) space. CL in the input space sort the sorts the training samples from the ``easier'' ones to ``harder'' ones\cite{luo2021medical,jimenez2019medical}. And CL in the output (task) space\cite{coarse-to-fine:2020} treats the original task as a ``harder'' task, and learns ``easier'' sub-tasks prior to learning the original ``harder'' task. In classification tasks with multiple ($\geq 3$ classes) classes, CL in task space creates ``easier'' classification sub-tasks out of the original ``harder'' task, usually by grouping classes with similarities\cite{coarse-to-fine:2020}, and thus reducing the number of total classes and the difficulty of the task.

In this work, we propose an end-to-end curriculum learning in task space strategy on classifying three categories of the Full-Field Digital Mammography (FFDM), namely \textit{Malignant}, \textit{Negative}, and \textit{False recall} (shown in Figure \ref{ffdmexamples}). Our method treats the original three-class classification task as a ``harder'' task in terms of CL, and create an ``easier'' sub-task of classifying the false recall cases against the combined group of the negative and the malignant cases. We introduce a loss scheduler to dynamically weight the contribution of the two tasks to the loss that the machine learning model learns from throughout the training process. We conduct experiments using an FFDM dataset of 1,709 images using 5-fold cross validation. The results show that our curriculum learning strategy can boost the model's performance compared to the baseline training strategies.

\section{METHOD}
\label{sec:method}  % \label{} allows reference to this section
\subsection{Study Cohort and Dataset}
In this IRB-approved study, we used a cohort of 1,709 FFDM images (349 Malignant cases, 653 Negative cases and 707 False recall cases). The FFDM images were retrospectively collected at our institution and reviewed by expert radiologists. Sample images from the three classes are shown in Figure \ref{ffdmexamples}. We conducted our experiments with 5-fold cross validation. Within each iteration of cross validation, one fold of the data was used as a hold-out test set. 80\% and 20\% of the remaining four folds were used as training and validation set respectively. The data partitioning scheme is described with more details in Table \ref{datadist}. Note that the numbers in Table \ref{datadist} are within $\pm1$ range, since the five folds do not evenly split the data samples.

\begin{table}[H]
    \caption{The number of images in each class for each partition of the 5-fold cross validation. Note that the numbers are within $\pm1$ range, since the five folds do not evenly split the data samples.}\label{datadist}
    \centering
    % \begin{tabular}{lM{4cm}M{2.5cm}M{2.5cm}M{2.5cm}}
    % \begin{tabular}{lc{4cm}c{2.5cm}c{2.5cm}c{2.5cm}}
    \begin{tabular}{lcccc}
    \toprule
     &  Malignant & Negative & False recall & Total\\ \midrule
    Training set &  226 & 425 & 460 & 1,111\\
    Validation set &  53 & 98 & 106 & 280\\
    Test set & 45 & 130 & 141 & 341\\
    Total & 349 & 653 & 707 & 1,709 \\\bottomrule
    \end{tabular}
\end{table}

\subsection{Curriculum in Task Space}
\label{cl-in-task}
Curriculum learning seeks a certain ``easy-to-hard'' order of the learning procedure. The order can be in terms of the input samples or the learning tasks. Here in this study, while we focus on the original three-class classification of the FFDM images, we treat it as a ``harder'' task and create an ``easier'' sub-task of binary classification. In this sub-task, the goal is to classify False recall against the combined group of Malignant and Negative cases.

Consider an image-label pair $(x_i, y_i)$, where $x_i$ represents the image and $y_i \in \{ 0,1,2\}$ (with $0$, $1$, and $2$ representing False recall, Negative, and Malignant respectively) is the ground truth label of the image. Let $f(x_i)^{(c)}$ be the probability that the predicted label is $c$ with $f(\cdot)$ being the machine learning model and $c \in \{ 0,1,2\}$. We compute the cross entropy loss for the original ``harder'' task of the three-class classification as in Equation (\ref{loss:3class}), where $y_{i,c}$ is the $c^{th}$ element in the one-hot representation of the label $y_i$.
\begin{equation}
    \La_{\text{hard}}( x_i, y_i ) = -\sum_{c} \left( y_{i,c} \cdot \log(f(x_i)^{(c)}) \right)
    \label{loss:3class}
\end{equation}

For the created ``easier'' task, to group the Negative and Malignant classes together against the False recall class, we assign $x_i$ a new label $z_i=\mathbbm{1}_{y_i \neq 0}$ where $\mathbbm{1}$ is the indicator function, which means that $z_i$ will only be $0$ if $y_i=0$, or else $z_i=1$. In terms of the prediction, we treat the probability of $z_i=0$ still as $f(x_i)^{(0)}$ and the probability of $z_i=1$ as $f(x_i)^{(1)} + f(x_i)^{(2)}$, or $1-f(x_i)^{(0)}$. Consequently, we can compute the binary cross entropy loss for the created ``easier'' task of the binary classification, as shown in Equation (\ref{loss:binary}).

\begin{equation}
    \La_{\text{easy}}( x_i, z_i ) = - \left( z_i \cdot \log(f(x_i)^{(0)})  + (1-z_i) \cdot \log(1-f(x_i)^{(0)}) \right)
    \label{loss:binary}
\end{equation}
\begin{equation}
    \La( x_i, y_i, z_i ) = \lambda \cdot \La_{\text{easy}}( x_i, z_i ) + (1-\lambda) \cdot \La_{\text{hard}}( x_i, y_i )
    \label{loss:combined}
\end{equation}

\begin{figure}[htb] 
    \begin{minipage}[b]{0.45\textwidth} 
        \centering 
        \includegraphics[width=0.8\textwidth]{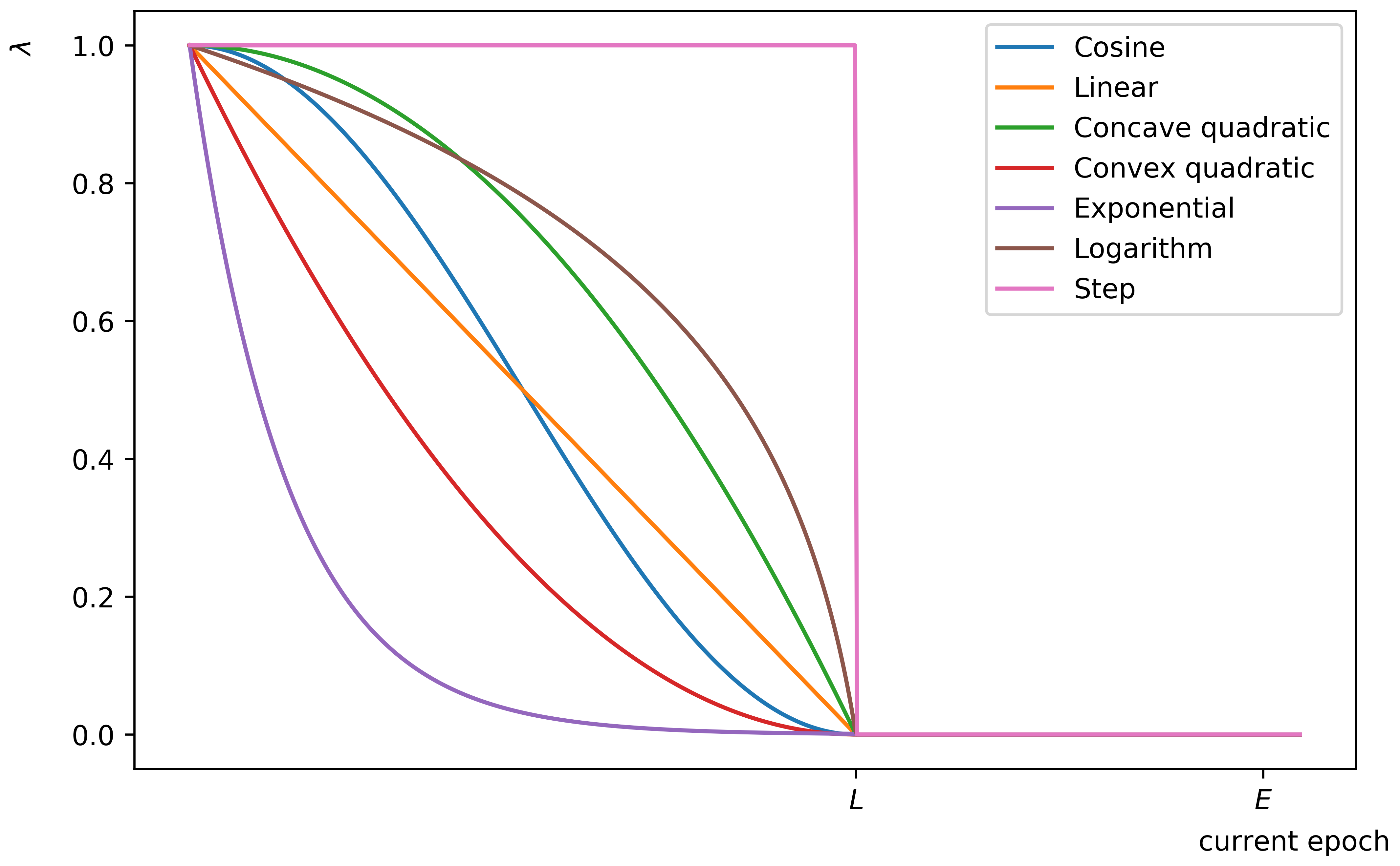} 
        \caption{Seven different types of loss scheduler. $L$ is a preset hyperparameter and $E$ is the total number of the training epochs} 
        \label{loss-scheduler-fig} 
    \end{minipage}% 
    \begin{minipage}[b]{0.55\textwidth} 
        \begin{table}[H]
        \caption{Functions of different types of loss scheduler with current epoch number, $e$, less than the preset hyperparameter $L$, ($\lambda=0$ when $L \leq e \leq E$)}\label{loss-scheduler-table} 
        \centering
        \begin{tabular}{ll} \toprule
            Loss scheduler type & Function (with $0 \leq e<L$)\\ \midrule
            Cosine    & $\lambda(e)=\left(\cos(e\pi / L)+1 \right) / 2$ \\ 
            Linear   & $\lambda(e)=1-e/L$ \\ 
            Concave quadratic & $\lambda(e)=-(e/L)^2+1$ \\ 
            Convex quadratic  & $\lambda(e)=L^{-2} \cdot (e-L)^2$ \\ 
            Exponential    & $\lambda(e)=\epsilon^{e/L}, \ \epsilon=10^{-3}$ \\ 
            Logarithm    & $\lambda(e)=\log(1+L-e) / \log(1+L)$ \\ 
            Step    & $\lambda(e)=1$ \\ \bottomrule
        \end{tabular} 
        \end{table}
    \end{minipage} 
\end{figure}

We compute the final loss as a weighted sum of $\La_1( x_i, y_i )$ and $\La_2( x_i, z_i )$ as shown in Equation (\ref{loss:combined}). We introduce a loss scheduler to dynamically weight the contribution of the losses from the two tasks throughout the entire training process, which is done by designing a specific function for the $\lambda$ in Equation (\ref{loss:combined}) with respect to the current training epoch number. In this work, we investigate the loss schedulers listed in Table \ref{loss-scheduler-table} and Figure \ref{loss-scheduler-fig}. In practice, $\lambda$ can be any function. Note that in Figure \ref{loss-scheduler-fig} and Table \ref{loss-scheduler-table} all loss schedulers have $\lambda=1$ at the beginning of training, and $\lambda=0$ after when the current epoch number, $e$, is larger than a preset hyperparameter $L$. This is to make the training focus on the ``easier'' task in the beginning of training, while transitioning to focusing on the ``harder'' task (original three-class classification) as training progresses.

\begin{table}[H]
    \caption{Comparison of the different methods' performance. ``LS: X'' stands for a certain type of loss scheduler. The bold numbers correspond to the highest value for each metric.} \label{results-table}
    \centering
    % \begin{tabular}{lM{4cm}M{2.5cm}M{2.5cm}M{2.5cm}}
    % \begin{tabular}{lc{4cm}c{2.5cm}c{2.5cm}c{2.5cm}}
    \begin{tabular}{lM{1.5cm}M{1.5cm}M{1.5cm}M{1.5cm}M{1.5cm}}
    \toprule
     &  Accuracy & Balanced accuracy & Average AUC & Binary task accuracy & Binary task AUC\\ \midrule
    Aboutalib et al.\cite{aboutalib2018deep} (baseline) & 0.489 & 0.458 & 0.658 & 0.598 & 0.633 \\\midrule
    LS: exponential & 0.491 & 0.468 & 0.658 & 0.623 & 0.641 \\
    LS: convex quadratic & 0.510 & 0.474 & \textbf{0.672} & 0.607 & 0.648 \\
    LS: linear & 0.480 & 0.476 & 0.655 & 0.614 & 0.635\\
    LS: cosine & 0.501 & 0.464 & 0.653 & 0.611 & 0.645\\
    LS: concave quadratic & 0.508 & 0.492 & 0.671 & \textbf{0.633} & 0.647 \\
    LS: logarithm & 0.511 & \textbf{0.494} & 0.668 & 0.617 & 0.644 \\
    LS: step & \textbf{0.515} & 0.483 & 0.669 & 0.622 & \textbf{0.655} \\\bottomrule
    \end{tabular}
\end{table}

\section{RESULTS}
We evaluate our method on 1,709 FFDM images using 5-fold cross validation. The results shown in Table \ref{results-table} are means over the five partitions. We use the VGG16\cite{simonyan2014very} as the backbone of the convolutional neural network (CNN) model. We present the results for different loss schedulers and compare our methods with Aboutalib et al.'s method\cite{aboutalib2018deep} (baseline) which is a deep learning method on the same three-class classification task as ours. Our model evaluation metrics include the accuracy and the AUC for the three-class classification, and the balanced accuracy by averaging the accuracies of the three classes, which reduces the effect induced by data imbalance. In addition, to further evaluate the methods' overall ability to distinguish False recall, we also show the accuracy and AUC of the binary ``easier'' task mentioned in Section \ref{cl-in-task}. All experiments are implemented in PyTorch framework and run on an Nvidia TESLA V100 GPU from Pittsburgh Supercomputing Center.

\section{DISCUSSION}
In this work, we proposed a curriculum learning strategy in task space for FFDM image classification of the three classes of Malignant, Negative, and False recall. We designed a loss scheduler to weight the contribution of the original ``harder'' three-class classification task and the created ``easier'' binary classification task.

According to the results, CL in task space with most forms of the loss scheduler outperform the baseline, which attributes to the fact that the loss schedulers focus more on the ``easier'' task during the first stage of the training process ($0\leq e < L$), while Aboutalib et al.'s method\cite{aboutalib2018deep} (baseline) only uses the basic learning strategy. In addition, the three concave loss schedulers, namely the concave quadratic, logarithm and step function, perform better than the other ones. This is partially because these three concave loss scheduler put reasonably more weight than others on the ``easier'' task, which provides more help on understanding the ``harder'' task (see Figure \ref{loss-scheduler-fig}).

\section{NEW OR BREAKTHROUGH WORK TO BE PRESENTED}
As far as we know, this work is one of the newest studies that develop a learning strategy with regard to curriculum learning in task space on FFDM image classification. Our method demonstrates that by scheduling the learning from an ``easier'' sub-task to the original ``harder'' task, the model's performance to classify the three classes of the FFDM images will be further improved from the baseline learning strategies. We believe that such curriculum learning methods will expand the possibilities of artificial intelligence (AI) studies and its applications in future medical imaging research.

% References
\bibliography{report} % bibliography data in report.bib
\bibliographystyle{spiebib} % makes bibtex use spiebib.bst

\end{document}